\documentclass[letterpaper, 10 pt, conference]{ieeeconf}  

\IEEEoverridecommandlockouts                              

\usepackage{amsmath} 
\usepackage{graphicx}
\usepackage{cleveref}
\usepackage{cite}
\usepackage{soul,xcolor}
\usepackage[acronym]{glossaries}
\usepackage{siunitx}
\usepackage{comment}

\newacronym{vs}{VS}{variable stiffness}
\newacronym{rom}{RoM}{range of motion}
\newacronym{fea}{FEA}{finite element analysis}

\title{\LARGE \bf JAMMit! Monolithic 3D-Printing of a Bead Jamming\\Soft Pneumatic Arm}
\author{Yao Yao$^{1}$, Maximilian Westermann$^{1}$, Marco Pontin$^{1}$, Alessandro Albini$^{1}$, and Perla Maiolino$^{1}$
\thanks{*This work was supported by  Engineering and Physical Sciences Research Council (EPSRC) Grant EP/V000748/1}
\thanks{$^{1}$Yao Yao, Maximilian Westerman, Marco Pontin, Alessandro Albini and Perla Maiolino are with Oxford Robotics Institute, University of Oxford, Oxford, OX1 2JD, United Kingdom;
{\tt\small yao.yao/maximilian.westermann/marco.pontin/}
{\tt\small alessandro.albini/perla.maiolino@eng.ox.ac.uk}}}

\begin{document}
\maketitle
\thispagestyle{empty}
\pagestyle{empty}

\begin{abstract}
3D-printed bellow soft pneumatic arms are widely adopted for their flexible design, ease of fabrication, and large deformation capabilities. However, their low stiffness limits their real-world applications. Although several methods exist to enhance the stiffness of soft actuators, many involve complex manufacturing processes not in line with modern goals of monolithic and automated additive manufacturing. With its simplicity, bead-jamming represents a simple and effective solution to these challenges. This work introduces a method for monolithic printing of a bellow soft pneumatic arm, integrating a tendon-driven central spine of bowl-shaped beads. We experimentally characterized the arm's range of motion in both unjammed and jammed states, as well as its stiffness under various actuation and jamming conditions. 
As a result, we provide an optimal jamming policy as a trade-off between preserving the range of motion and maximizing stiffness. The proposed design was further demonstrated in a switch-toggling task, showing its potential for practical applications.

\end{abstract}

\section{INTRODUCTION}
\begin{figure}[!t]
    \centering
    \includegraphics[width=0.95\columnwidth]{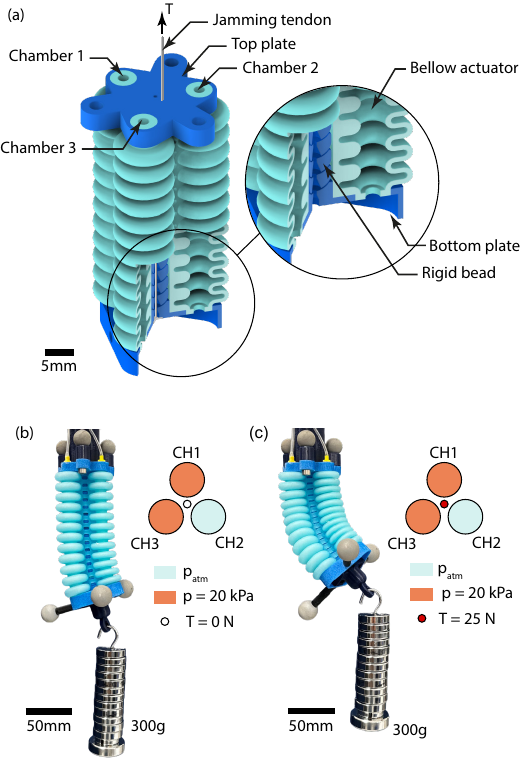}
    \caption{The variable stiffness soft arm. (a) The design of the soft arm enables monolithic fabrication and variable stiffness through beaded string jamming. (b) Performance of the soft arm while inflated, but without jamming enabled. (c) By engaging the integrated bead jamming mechanism, the load bearing capability of the arm drastically improves.}
    \label{fig:figure1}
\end{figure}

Soft robots offer a key advantage over traditional rigid robots - their inherent compliance allows safe interaction with unstructured environments and enables adaptation to complex tasks~\cite{softrobot1, softrobot2}. 
Pneumatic actuators play a central role in this domain due to their lightweight construction, high power-to-weight ratio, and ease of manufacturing and actuation ~\cite{xavier2022soft}. 
Recent advancements in 3D printing have enabled the automation of the fabrication of complex actuators and integrated systems, significantly reducing production time and cost while improving precision and repeatability~\cite{3dprinting, wang2022modular, zhai2023desktop, wehner2016integrated}.
However, the selection of commercial soft materials remains limited~\cite{3dprintingmaterial}, with elongation at break and actuation pressures significantly lower than traditional silicones, restricting load-bearing capacities and overall stiffness.

To address these limitations, designs such as pleated or bellow actuators have been developed, using macroscopic geometric deformations rather than material strain to achieve desired performance~\cite{pleated/bellow}. While effective at enabling high deformations at low pressures, these actuators exhibit low stiffness, limiting their ability to exert significant forces or support heavier loads. This constraint restricts their application in tasks requiring both strength and flexibility.

A promising approach to overcoming this challenge are variable stiffness mechanisms~\cite{stiffness_review}. Existing methods, such as tendon-based or pneumatic antagonistic actuation, often require complex control and scale poorly for high degrees-of-freedom systems~\cite{tendon_stiffening, pneumatic_stiffening}.
Hybrid soft-rigid designs and jamming-based techniques (e.g., granular or fiber jamming) have also been explored~\cite{soft-rigid,selective_jamming, Strings_of_Beads, Jammkle, Jeg}. 
While these solutions enable stiffness modulation, they often involve complex assembly steps, which prevent monolithic fabrication, require airtight chambers for jamming media, and can lead to uneven stiffness distribution due to media settling under external forces~\cite{VSspine}.

In this paper, we propose a novel variable stiffness soft pneumatic arm design based on bead jamming. The proposed design, shown in \cref{fig:figure1}a, is monolithically 3D-printed, integrating a central channel for a tendon that applies jamming tension. This design minimizes manufacturing steps while achieving effective stiffness modulation. The spherical beads in the central chain allow omnidirectional motion and maintain \gls{rom} comparable to other soft arms. Experimental results demonstrate significant stiffness enhancement under load, as indicated in \cref{fig:figure1}b and c, by reduced sag when jamming is applied. The remainder of the paper is structured as follows. Section \ref{sec:methods} details the design and manufacturing process. Section \ref{sec:setup} outlines the experimental setup, and Section \ref{sec:exp_descr} presents validation experiments and results. Conclusion follows.

\section{Materials and Methods}
\label{sec:methods}
\subsection{Soft Arm Design and Manufacturing}
\begin{figure}
    \centering
    \includegraphics[width=0.9\columnwidth]{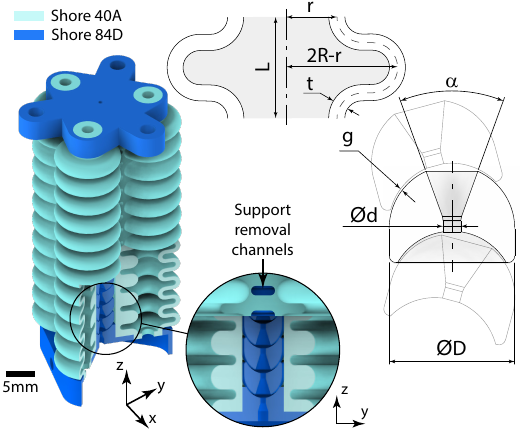}
    \caption{Materials and geometry of the variable stiffness soft arm. Soft regions are printed with 40A Shore hardness, while rigid ones reach 83-86D Shore hardness. A clearance $g$ of 0.2\,mm allows for the printing of the beads without them fusing together, while openings on the side walls between the bellow actuators enable easier removal of the support material during post-processing of the print. 
    }
    \label{fig:man}
\end{figure}
As displayed in \cref{fig:man}, the soft bellows are printed on a Stratasys J735 \textsuperscript{\textregistered} (Stratasys Ltd, USA) multi-material 3D printer using a digital material blend of VeroCyan\textsuperscript{\textregistered} (Stratasys Ltd, USA) and a rubber-like soft material Agilus30\textsuperscript{\textregistered} (Stratasys Ltd, USA) to achieve a Shore hardness of 40A, while the end plates and the beads are fabricated using a rigid plastic-like material VeroCyan\textsuperscript{\textregistered} (Stratasys Ltd, USA) with a Shore hardness of 83-86D. Soluble support material SUP706\textsuperscript{\textregistered} (Stratasys Ltd, USA) was also used during the print for the voids.

The base geometry for the bellow actuators is the result of an optimization process our group developed in \cite{bellow_actuator}. Three bellows are joined together at 120$^\circ$ to form the soft arm and a further optimization process employing \gls{fea} was conducted to get to a geometry of the soft arm that can achieve 90$^\circ$ omnidirectional bending at 20\,kPa \cite{bellowdesign2}. The resulting geometrical parameters are as follows: inner radius $r$=4\,mm, wall thickness $t$=1.5\,mm, average radius $R$=7\,mm, module length $L$=7.2\,mm, and module number $12$ for each bellow pneumatic chamber. These constraints allowed for a central column with a diameter of 10\,mm to be cut out, to house the beads for the bead jamming mechanism.

The design of the bowl-shaped beads results from the integration of a pre-existing design ~\cite{bead_design} with our monolithic manufacturing goal. With respect to \cref{fig:man}, large bead diameter $D$ maximizes the holding torque of the beaded-string for a given jamming tension \cite{bead_design}. A minimum clearance $g$ between each bead and between the beads and the central column wall must be granted for the inclusion of support material, to avoid the various elements of the design fusing together while printing. Whilst this clearance is mandatory, minimizing it is crucial for effective jamming performance, as this minimizes tendon travel to engage the beads. Clearance test prints were performed to empirically evaluate the 0.2\,mm minimum reliable clearance $g$ achievable with the printer used. Given these considerations, the beads were designed with an external diameter $D$ of 9.2\,mm. Each bead features a central hole with diameter $d$ of 1.3\,mm, for the 1\,mm thick nylon jamming tendon, while the angle $\alpha$ of of $30^{\circ}$ coupled with the spherical joint resulting from the bead design ensures unrestricted range of motion of the soft arm. The dimensions allowed for 17 beads to be included in the soft arm, with the top and bottom ones fused into the soft arm's end plates.

To enable the effective removal of the support material, openings were designed in correspondence of each bead pair as show in the detailed view of \cref{fig:man}. Openings at the inlet and outlet of the soft bellows are also present for the same reason. During post processing, a steel wire was run through the central tendon channel, to remove the support material. A pressurized water jet was then flushed through the tendon channel and this forced most of the support material in and in between the beads to be expelled through the side openings. The soft arm was finally put in a solution of $0.02\,$kg/L Sodium Hydroxide and $0.01\,$kg/L Sodium Metasilicate at 30\,$^\circ\mathrm{C}$ for 3 days, with additional manual cleaning under warm water every day, for the removal of the remaining support material. Once clean, bellows were sealed at the bottom with high-viscosity superglue (UN3334 Everbuild Building Products, UK), while tubing was applied at the inlets to enable their pressurization. The nylon jamming tendon was routed last, knotting its end to allow it to compress the beads through the bottom plate.

\subsection{Finite Element Analyis of the Beads}
\begin{figure}
    \centering
    \includegraphics[width=0.9\columnwidth]{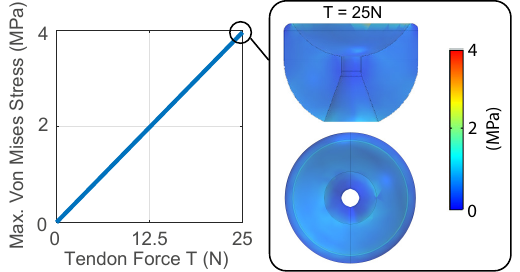}
    \caption{The FEM results for determining the maximum tension force: (left) the maximum volumetric von Mises stress as a function of the applied force; (right) the volumetric maximum von Mises stress distribution of an intermediate bead, shown from a cross-sectional view through the center for T=25\,N which is used as the upper limit for the tension applied to the real system. 
    }
    \label{fig:FEM}
\end{figure}
To determine the maximum allowable tension force for the jamming tendon, a \gls{fea} was employed using COMSOL Multiphysics\textsuperscript{\textregistered} (COMSOL Inc., Sweden). As mentioned, the beads are fabricated using the rigid material VeroCyan, with a Young's modulus ranging from 2000 to 3000\,MPa, tensile strength between 50 and 65\,MPa, and density between 1.17 and 1.18\,g/cm³. Therefore, in the \gls{fea}, a linear elastic material model was used, with a representative Young’s modulus of 2500\,MPa and a density of 1.175\,g/cm³, alongside a Poisson’s ratio of 0.38~\cite{vero_poisson_ratio}. During the simulation, the beads were arranged in a vertical stack, replicating their configuration in the soft arm at rest. The uppermost bead was fixed in place, and an upward body force was incrementally applied to the bottom-most bead, ranging from 0\,N to 25\,N, with a step size of 5\,N. The contact interfaces between adjacent beads were modelled using identity pairs to replicate the no-slip condition between the beads when jammed. The maximum jamming force was limited to 25\,N to ensure a safety factor of 12 given the bead material tensile strength of 50\,MPa. This was done to guarantee the system's long-term operational integrity and account for mechanical fatigue and uncertainties in material properties.
\Cref{fig:FEM} shows the maximum volumetric Von Mises stress as a function of the applied jamming force.

\section{Experiments and Results}
\label{sec:exp_descr}

The experiments aim at 
characterizing the soft arm \gls{rom}, in both jammed and unjammed states, and evaluating the resulting sag 
under load at different actuation pressures and jamming conditions. 

\subsection{Experimental Setup}
\label{sec:setup}

\begin{figure}
    \centering
    \includegraphics[width=0.9\columnwidth]{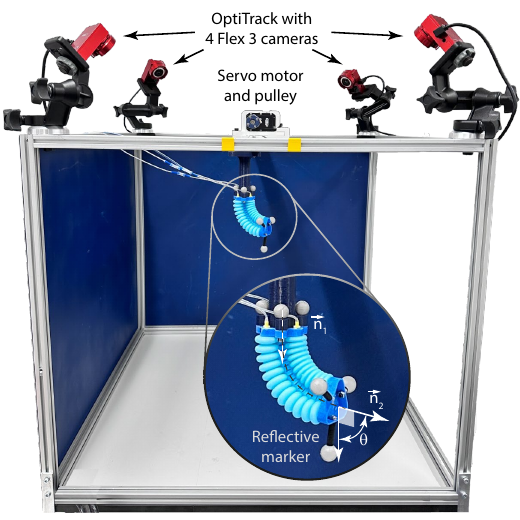}
    \caption{The setup includes a tracking system with four cameras mounted on a $780\ \mathrm{mm} \times 700\ \mathrm{mm} \times \ 563\ \mathrm{mm}$ aluminum frame, reflective markers on the soft arm’s top and bottom plates for angle measurement, a servo for tendon tensioning.}
    \label{fig:setup}
\end{figure}

As illustrated in~\cref{fig:setup}, the experimental setup utilized a Motive OptiTrack\textsuperscript{\texttrademark} system (NaturalPoint, Inc., USA) with four Flex 3 cameras to track the positions and orientations of the soft arm, mounted on a $780\ \mathrm{mm} \times 700\ \mathrm{mm} \times \ 563\ \mathrm{mm}$ aluminum frame. 
Three reflective markers were placed for traking purpose on both end plates of the soft arm.
The geometric centre of the three makers defines the position of the base and tip of the arm in Cartesian space.
The rotation angle $\theta$, as indicated in \cref{fig:setup}, was computed as the angle between the normal vectors of these two rigid bodies. 

A DYNAMIXEL MX-106R Servo Motor \textsuperscript{\textregistered} (Robotis, UK) was mounted above the soft arm to actuate the jamming tendon through a pulley system. An Anest Iwata\textsuperscript{\textregistered} air compressor (Anest Iwata Corporation, Japan) supplied compressed air to a VPPE-3-1-1/8-2-010-E1 Festo\textsuperscript{\textregistered} proportional-pressure regulator (Festo AG \& Co. KG, Germany), used to control the actuation pressure.

\subsection{Range of Motion Characterization}
\label{sec:charact}

The purpose of this characterization is to evaluate the \gls{rom} of the soft arm under unjammed and jammed conditions.
\begin{figure}[]
    \centering
    \includegraphics[width=0.9\columnwidth]{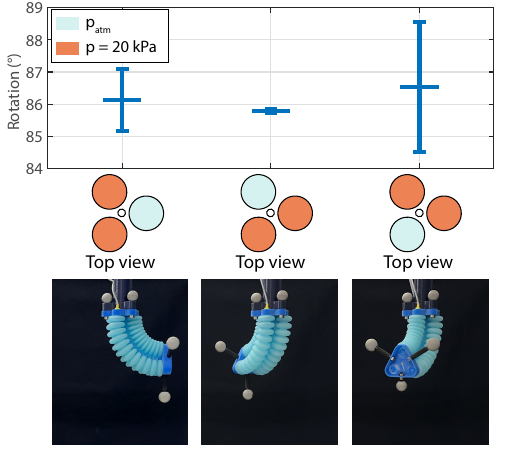}
    \caption{Range of motion and uniform deformation test: rotation angle results for each direction, with two pneumatic chambers actuated simultaneously at 20 kPa.}
    \label{fig:threedir}
    \vspace{-2em}
\end{figure}

Before comparing these two conditions, a preliminary test was conducted to assess whether the arm deforms uniformly in different directions. In this test, the arm was actuated in the three principal directions of motion by simultaneously actuating the pneumatic chambers in pairs at 20\,kPa, maximum pressure the bellow actuators could withstand without permanent damage due to excessive material stretch. 

Data was collected at a steady state during the final $2$ seconds after allowing the system to settle for at least $15$ seconds following pressure application. For each pair of actuated chambers, the test was repeated three times. The averaged rotation angles and the corresponding standard deviations are displayed in \cref{fig:threedir}. The results show that the rotation angles in the three directions are $86.13^{\circ}$, $85.78^{\circ}$, and $86.32^{\circ}$, with standard deviations of $0.96^{\circ}$, $0.07^{\circ}$, and $2.03^{\circ}$, indicating good repeatability. A one-way ANOVA test performed on the experimental results showed no statistically significant difference between the three directions.

The characterization of the effect of jamming on the \gls{rom} then followed.
To begin with, three different actuation sequences were considered:  
\begin{itemize}
    \item \textbf{Sequence 1}: The arm was first actuated to 20 kPa, followed by the application of the full tension force of 25 N. 
    \item \textbf{Sequence 2}: A pre-tension of approximately 0.48 N was applied to engage the beads before actuating the arm at 20 kPa, followed by the full tension force. 
    The pre-tension force represents the minimum empirically determined force, resulting from the servo motor’s resolution (0.4 N) and the friction present in the system.
    \item \textbf{Sequence 3}: The arm was first actuated to 20 kPa, followed by the application of pre-tension, and finally the full tension force.
\end{itemize}
These three sequences were selected to assess the effect of the 0.2\,mm inter-bead gap and the jamming tension transient on the stiffening behavior of the arm. Sequence 2 in particular was developed to evaluate the \gls{rom} of the soft arm when the gap was removed through the application pre-tension force.

\begin{figure}[]
    \centering
    \includegraphics[width=0.9\columnwidth]{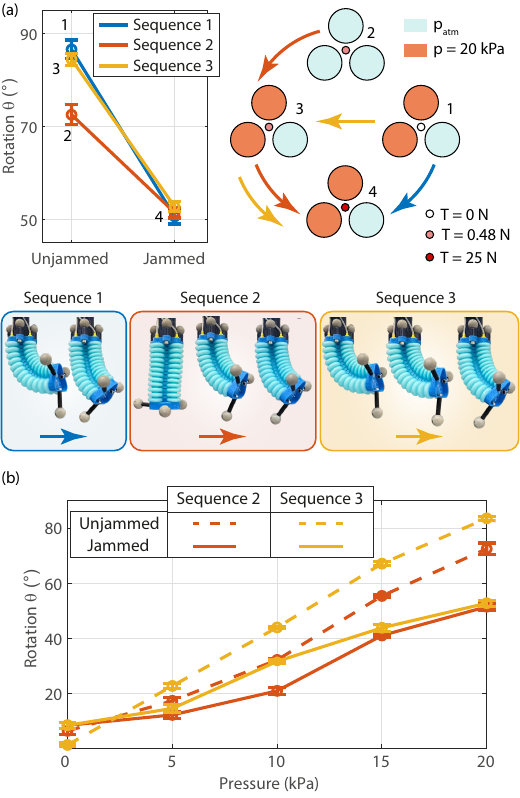}
    \caption{Range of motion test. (a) Preliminary results for three actuation-jamming sequences: The upper left plot shows the unjammed and jammed results for each sequence, while the upper right provides a schematic highlighting the actuation and tension force conditions for each sequence. The bottom row presents real photos of the soft arm demonstrating the status for each sequence. (b) Further range-of-motion characterization showing the unjammed and jammed rotation angles for Sequences 2 and 3 as pressure changes.}
    \label{fig:RoM}
    \vspace{-2em}
\end{figure}

During each sequence, the bending angle after pressurization and the one after full jamming were recorded through the OptiTrack system. The results of the experiment are shown in \cref{fig:RoM}a for each sequence, in terms of the rotation angle $\theta$. This was computed, as shown in \cref{fig:setup}, as the angle between the vectors $\overrightarrow{n_1}$ and $\overrightarrow{n_2}$. $\overrightarrow{n_1}$ is orthogonal to the plane defined by the markers of the base plate, while $\overrightarrow{n_2}$ is orthogonal to the plane defined by the markers at the tip of the arm. During the test, we noticed that the application of the jamming tension caused the rotation angle of the soft arm to change and $\overrightarrow{n_2}$ to drift out-of-plane. In the presentation and analysis of the results, we decided to neglect the out-of-plane rotation of the arm as this was an order of magnitude smaller than the in-plane one (e.g. $3.66^{\circ}$ compared to $32.29^{\circ}$ for sequence 3).

For each experimental trial, we allowed at least 15 seconds after each applied action for the system to reach a steady state. Data was then averaged over the final 2 seconds of this period to generate the corresponding data points. We then computed the mean rotation angle of the arm before and after jamming for each sequence across all trials, along with their standard deviations for comparison. 
The initial rotation angles before jamming for Sequences 1 and 3 were $86.53^\circ$ and $84.39^\circ$, respectively. The small difference between them is due to the accuracy of the pressure regulator, which has a minumum controllable pressure of 1kPa~\cite{festo_vppe_datasheet}. Sequence 2 exhibited the smallest initial rotation angle before jamming at $72.48^\circ$.
This result is to be expected, as Sequence 2 starts from a lightly jammed configuration, where the increased internal friction between beads reduces the bending angle $\theta$ compared to the other two initial unjammed states. Despite this difference, the final rotation angle after the soft arm is fully jammed are on average very similar ($50.60^{\circ}$, $51.42^{\circ}$, $52.69^{\circ}$ for sequences 1,2 and 3 respectively), with no statistically significant difference between the three (one-way ANOVA test with p-value=0.224).

From these observations, we can conclude that, when unjammed, the arm benefited from a large \gls{rom}, while, a noticeable reduction of the bending angle occurred when jamming was enabled. 

Due to the similar resulting behavior between Sequences 1 and 3, only Sequences 2 and 3 were utilized in the remainder of \gls{rom} characterization as they respectively led to the smallest and the largest average variation in the angle $\theta$ after jamming. \Cref{fig:RoM}b reports the value of $\theta$ at increasing levels of actuation pressures, for both Sequence 2 and 3.
The actuation pressure ranged from $0$ to 20\,kPa with 5\,kPa intervals, and with the full tension force of 25\,N applied. Three trials were conducted for each pressure level, and the average rotation angle and its standard deviation across trials were computed. Overall, the results starting from 5\,kPa indicate that Sequence 2 experienced a smaller reduction in rotation angle after jamming, while Sequence 3 maintained a larger jammed rotation angle across the pressure range, consistent with the preliminary findings. For the unjammed results, the pressure-only condition (yellow dashed line, Sequence 3) shows a steady increase in rotation angle as pressure increases. The pre-tension condition followed by actuation pressure (red dashed line, Sequence 2) demonstrates a similar trend but with slightly lower angles starting from 5 kPa. Finally, the jammed results (solid lines) for both sequences reveal a more substantial decrease in rotation angle compared to the unjammed results of Sequence 2, suggesting a stronger effect of full-tension over pre-tension. From the analysis, Sequence 3 represents a better choice with respect to the other two as it allows for a slightly larger \gls{rom}.

It is worth noting that when no pressure and no tension were applied (unjammed results of Sequence 3), the rotation angle was $1.21^{\circ}$. This angle increased to $6.31^{\circ}$ when pre-tension was applied (unjammed results of Sequence 2), and further increased to $8.33^{\circ}$ with full tension in both sequences (jammed results of both sequences). This increase results from the inherent asymmetry of the three bellow chambers, as well as gaps and slight misalignments between beads, leading to an unavoidable change in orientation when the tendon was pulled upwards to fully engage the beads. 

\subsection{Variable Stiffness Validation}

\begin{figure}[t]
    \centering
    \includegraphics[width=0.95\columnwidth]{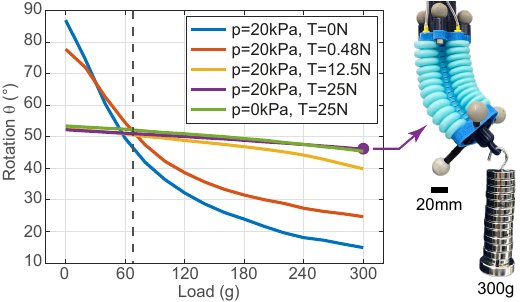}
    \caption{Stiffness characterization results: the change in rotation angle with incremental loads under different actuation and jamming conditions. The dashed line in the chart defines two regions based on load range to indicate when jamming (to the right of the line) or no jamming (to the left of the line) is preferable concerning the soft arm’s sag.}
    \label{fig:stiffness}
    \vspace{-1.2em}
\end{figure}

The \gls{vs} capability of the arm was evaluated by measuring the sag occurring when a load was applied to the robot's tip. Experiments were performed with weights ranging from $0$ to 300\,g (equal to 3.3 times the weight of the soft arm at 90.1\,g), in increments of 20\,g. 

Furthermore, the sag was evaluated under actuations and jamming conditions corresponding to Sequence 3:
\begin{itemize}
    \item Pressure-only actuation.
    \item Pressurization, followed by 0.48\,N jamming tension.
    \item Pressurization, followed by 0.48\,N and finally 12.5\,N jamming tension.
    \item Pressurization, followed by 0.48\,N and finally 25\,N jamming tension.
    \item Pressurization, followed by 0.48\,N and 25\,N jamming tension and finally depressurization.
\end{itemize}
Similarly to the \gls{rom} characterization tests, after applying the jamming/actuation sequence, we let the system reach the steady state before applying the loads to the tip of the soft arm. The measured position of the tip was recorded after settling and averaged considering a time window of 2\,s.

\Cref{fig:stiffness} illustrates the change in bending angle as a function of the external load ~\cite{force-angle1,force-angle2} for different jamming conditions.
As expected, when no jamming was applied, the system reached the highest initial bending angle of 86.99$^{\circ}$, and the angle decreased dramatically even for the lower weights. A similar trend was observed when a low jamming tension of 0.48\,N was applied to the system after the initial actuation: the initial angle of 77.77$^{\circ}$ decreased rapidly and finally reached 24.70$^{\circ}$. The remaining three conditions demonstrated comparable performance, with rotation angles starting at 52.6$^{\circ}$, 52.24$^{\circ}$, and 53.46$^{\circ}$, and ending at 39.85$^{\circ}$, 46.04$^{\circ}$, and 45.4$^{\circ}$, with the lower angle achieved with the smaller jamming tension of 12.5\,N. 

As shown in \cref{fig:stiffness}, two distinct regions can be identified, separated by the vertical dashed line. When the external load is below 70\,g (region to the left of the line) one would get a larger sag by jamming the arm compared to supporting the load without jamming. Unjammed operation should therefore be preferred. Jamming becomes advantageous for loads higher than 70\,g, where it successfully manages to drastically reduce the sagging of the robot arm tip (region to the right of the line). 

\subsection{Demonstration in a Switch-Toggling Task}
\begin{figure}[t]
    \centering
    \includegraphics[width=0.95\columnwidth]{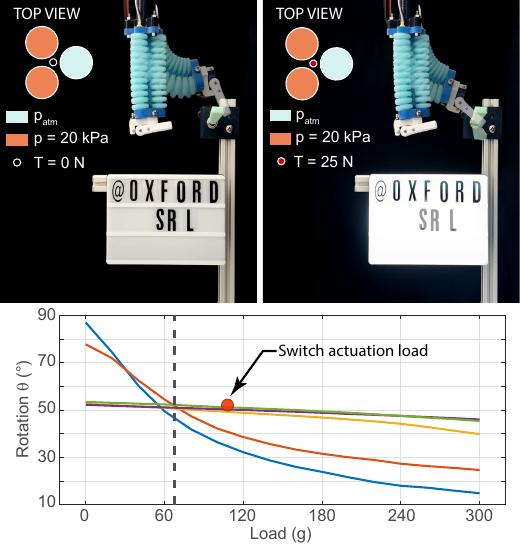}
    \caption{Experimental validation in a switch-toggling task. The top part shows the arm's behavior under unjammed (left) and jammed (right) conditions. The bottom plot replicates Figure 6, with the switch actuation load (110\,g) indicated. The plot demonstrates that the arm in the unjammed state cannot apply sufficient force to toggle the switch, while the jammed state increases stiffness and ensures successful task completion.}
    \label{fig:task}
\end{figure}
The effectiveness of the proposed bead jamming mechanism in increasing the force application capabilities of the soft arm was validated in a task where the robot was controlled to toggle a lever switch.

A SG-90 servo motor, capable of providing 2.5\,kg$\cdot$cm of torque, served as the end effector and was mounted on the tip of the soft arm, with a 4\,cm servo horn connected to its output shaft. A lever switch with an 80\,g actuation load was wired to an off-the-shelf light box. The arm was pressurized at 20\,kPa and the servo motor was then actuated to toggle the switch. Two trials were performed, one with the arm in an unjammed state and the second with jamming activated. The results are displayed in \cref{fig:task}. When the soft arm was driven solely by an actuation pressure of 20\,kPa, the task failed as the arm was pushed up by the lowering servo horn. In the second scenario, instead, where a jamming tension of 25\,N was applied following the initial pressurization,  the switch was successfully toggled, activating the light box. A detailed demonstration is provided in the supplementary video.

\section{Conclusion}
\label{sec:conclusions}
In this paper we proposed a monolithically 3D printed bellow soft arm, integrating a \gls{vs} mechanism based on bead-jamming. The spherical beads were designed as to not interfere with the large native \gls{rom} (approximately \SI{90}{\degree}) when unjammed.

Experimental characterization shows that the proposed solution notably increases stiffness, reducing the sag of the arm in the presence of external loads. However, as a side effect, jamming also reduces the \gls{rom} of the arm. In this regard, we found that the actuation/jamming sequence affects the mobility of the actuator. To investigate this, we evaluated the effect of three different actuation/jamming sequences and selected the one with the least impact on the overall \gls{rom}.

Future work will focus on closed-loop control, allowing real-time stiffness adjustment to adapt the actuator’s behavior to task or environmental changes.

\addtolength{\textheight}{-12cm}   





\section*{ACKNOWLEDGMENT}

The authors thank Mr. Giammarco Caroleo (University of Oxford) for his contributions in shaping the research goals.


\bibliographystyle{ieeetr}
\bibliography{ref}

\end{document}